\documentclass[11pt,a4paper]{article}
\usepackage[hyperref]{emnlp2018}
\usepackage{times}
\usepackage{latexsym}

\usepackage{url}
\usepackage{amsmath}
\usepackage{graphicx}
\usepackage{subfigure}
\usepackage{bm}

\aclfinalcopy 

\title{Exploration on Grounded Word Embedding: \\
Matching Words and Images with Image-Enhanced Skip-Gram Model}

\author{Ruixuan Luo \\
  MOE Key Laboratory of Computational Linguistics, School of EECS, Peking University \\
  {\tt luoruixuan97@pku.edu.cn} }

\date{}

\begin{document}
\maketitle
\begin{abstract}
Word embedding is designed to represent the semantic meaning of a word with low dimensional vectors. The state-of-the-art methods of learning word embeddings (word2vec and GloVe) only use the word co-occurrence information. The learned embeddings are real number vectors, which are obscure to human. In this paper, we propose an Image-Enhanced Skip-Gram Model to learn grounded word embeddings by representing the word vectors in the same hyper-plane with image vectors. 
Experiments show that the image vectors and word embeddings learned by our model are highly correlated, which indicates that our model is able to provide a vivid image-based explanation to the word embeddings.
\end{abstract}

\setlength{\abovedisplayskip}{3pt}
\setlength{\belowdisplayskip}{3pt}
\setlength{\abovedisplayshortskip}{3pt}
\setlength{\belowdisplayshortskip}{3pt}

\section{Introduction}
Pre-trained word representations are a key component in many natural language understanding models. Different from the binary one-hot vector corresponding to a dictionary position, word embeddings are expected to map words with similar meanings to nearby points in a vector space. However, learning high quality representations can be challenging. Previous work focuses on learning the word representations via language modeling \citep{word2vec} or matrix factorization \citep{pennington2014glove}. The learned vectors are in the form of a sequence of real numbers, which are hard to explain. \citet{Vilnis2014Word,Athiwaratkun2017Multimodal} proposed to model each word by Gaussian distribution, but the assumed Gaussian distribution can be problematic and the meaning of the vectors still remain obscure to human beings. While many prior work learns word embeddings at the input level, there are also various methods to implicitly learn label/word embedding at the output levels \cite{sun2017label,wei2018,ma2018query}.

Images are natural resources that human can easily understand. With the help of vision, multi-modality is becoming a very important way to help understand natural language. Image caption generation has attracted much attention \citep{Xu2015Show,7298935,DBLP:journals/corr/abs-1808-08732}, which aims to automatically generate textual captions based on an image. \citet{malinowski2014multi,malinowski2014towards,malinowski2015ask,ren2015exploring,ma2016learning} focus on image based question answering. 
\citet{chen2018zero,gella2017image} proposed to use images as an intermediary between two languages in machine translation, based on the assumption that both the source language and target language are describing the same picture. 
Another research line that is related to our work is bidirectional image and sentence retrieval \citep{klein2015associating,socher2014grounded,ma2015multimodal,wang2016learning}. 

Inspired by the work of \citet{ijcai2017-438} that embodies image information into the representation of the triples in the knowledge graph, we propose to represent the word embeddings in the same hyper-plane with the vectors of images. This way, the word embeddings are connected with the images by embodying the information of images into the word vectors. Although the word embeddings alone are still not understandable to human beings, the images would serve as a more precise and vivid explanation. 

ResNet \citep{He2015Deep} has achieved great success in the task of image classification \citep{ILSVRC15}. In this paper, we apply a pre-trained ResNet to compress each image into a  vector. We train an image projection matrix to project the image vector to the same space of the word embeddings by predicting one  nearby word iteratively based on the central word and the image. An energy based objective function is applied to train the model.   

Our contributions lie in the following aspects,
\begin{itemize}
\item We propose to learn the grounded word embeddings by projecting them to the same hyper-plane with image vectors, which makes the word embeddings explainable to human beings.
\item We propose an Image-Enhanced Skip-Gram Model to jointly learn the image vectors and word embeddings. Experiment results show that the target image vectors and word embeddings are highly correlated.
\end{itemize}

\section{Image-enhanced Skip-gram Model}
\begin{figure}
\centering
\includegraphics[width=3in]{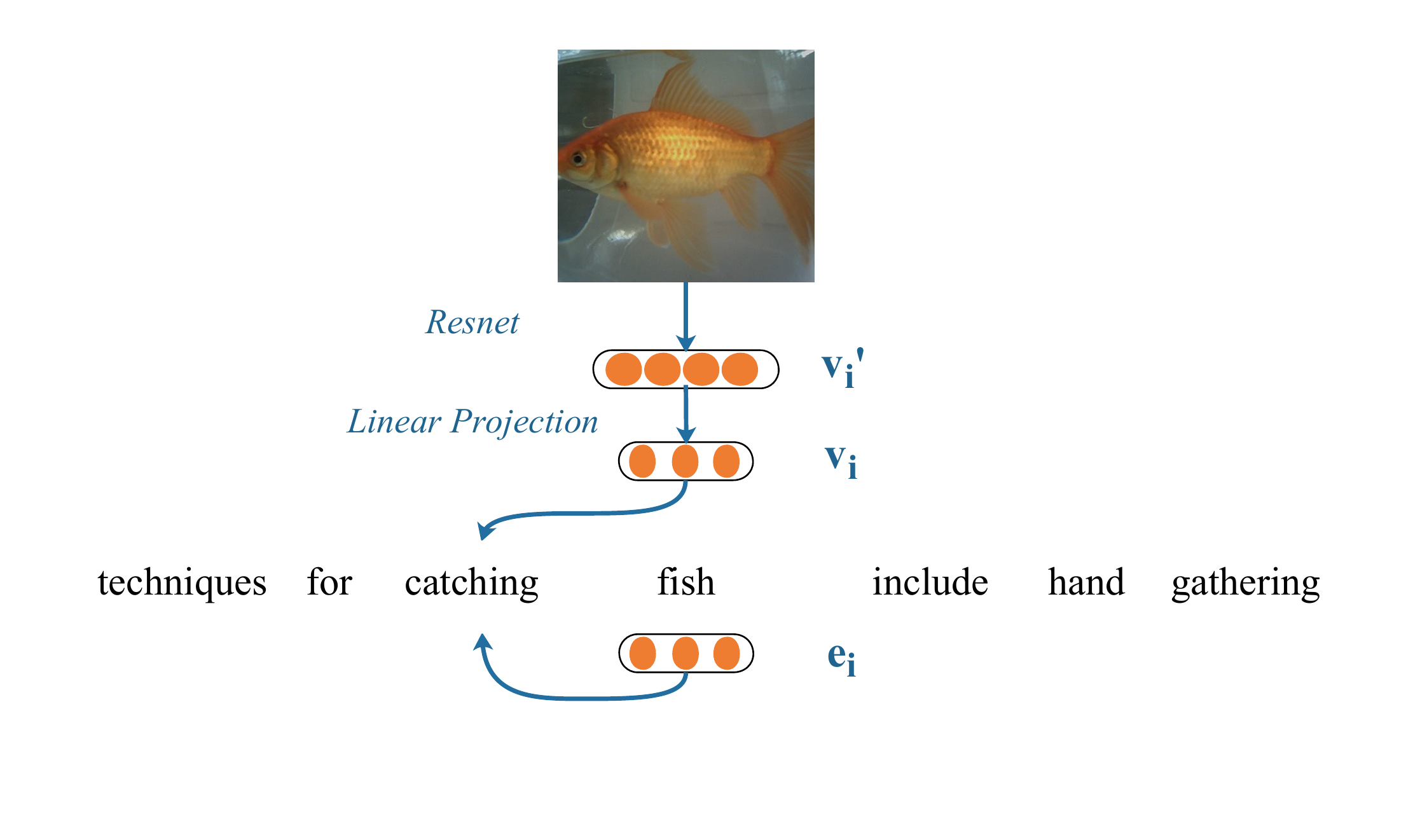}
\caption{An illustration of our Image-enhanced Skip-gram Model. ``Fish'' is the central word, and there is a corresponding image. Other words are the context words.}
\label{fig:model}
\end{figure}
\citet{word2vec} proposed the Skip-Gram Model to learn word embeddings. The intuition behind this model is to maximize the probability of observing a word $w_i$ given another nearby word $w_{i-j}$ within a window. This procedure follows the  hypothesis that words occurring in natural contexts tend to be semantically related. This model is both efficient and effective in capturing the semantic meaning behind the words, so we build our model based on the Skip-Gram Model.

Assume in a window of words, the central word $w_i$ has one corresponding image $f_i$, $w_c$ is a context word near to $w_i$, $w_c'$ is a negatively sampled word that is not in the window. We iteratively train the word embedding of $w_i$ and the vector of the image $f_i$ with the loss function with Equation \ref{eqn:word loss} and Equation \ref{eqn:figure loss} respectively. This way, the word embeddings and image vectors will gradually be distributed across the same space,
\begin{eqnarray}
\begin{split}
L(w_i,w_c,w_c') = \max(0,\ & m - \lambda E_{\theta}(w_i, w_c)  \\
				& + E_{\theta}(w_i, w_c')) \label{eqn:word loss}
\end{split}
\\
\begin{split}
L(f_i,w_c,w_c') = \max(0,\ & m - \lambda E_{\theta}(f_i, w_c) \\ 
				& + E_{\theta}(f_i, w_c')) \label{eqn:figure loss} 
\end{split}
\end{eqnarray}
where $E_{\theta}$ is the energy based function. $\lambda$ is a hyper-parameter that tunes the ratio between positive samples and negative samples.

We apply ResNet \citep{He2015Deep} to extract the vector of the image $v'_i$, which is then projected to a vector $v_i$ with the same dimension as the word embeddings by a projection matrix $W_p$,
\begin{align}
v_i' = & \text{ResNet}(f_i) \label{eqn:resnet} \\
v_i = & v_i' W_p 
\end{align}
To make the model more efficient, the parameters of the ResNet are fixed, the word embeddings are pre-trained with skip-gram model, while the word embeddings remain learnable during the training of IESG model.
The energy function is defined as the inner product between two vectors. For a word $w$, we use its word embedding $e$. For an image $f$, we use its image vector $v$,
\begin{align}
E_{\theta}(w_i, w_c) = & \ e_i \cdot e_c  \\
E_{\theta}(f_i, w_c) = & \ v_i \cdot e_c 
\end{align}
In Figure \ref{fig:model} we show an example, ``fish'' is the central word in the context of ``techniques for catching fish include hand gathering''. The window size here is 4 on both sides including the central word. We have an image corresponding to the word ``fish''. We convert the image $f_i$ into a vector $v_i'$ (Equation \ref{eqn:resnet}), then we use a linear layer to project $v_i'$ into $v_i$, which has the same dimension as word embeddings. We iteratively minimize $L(w_i,w_c,w_c')$ and $L(f_i,w_c,w_c')$. This way, both the image vectors and the word embeddings will be pushed to the same space. 

\subsection{Training Details}
To accelerate the convergence speed of the model, we first pre-train the word embeddings with the word2vec toolkit, the embeddings are not fixed. For the image part, we pre-train ResNet on the images in the image classification task (ImageNet), the parameters of the ResNet are then fixed. We use  stochastic gradient descent with momentum (Momentum SGD) to minimize the loss function, the learning rate is 0.01, the momentum is 0.9. The batch size is 16. The window of context is set to 5. The hyper-parameter $\lambda$ is 5. During training, we randomly select one of the images of a word. 

\section{Experiment}
We use the bidirectional image-word matching task to evaluate the learned word embeddings. 

\subsection{Dataset}
In this paper, we use the first 100,000,000 bytes of cleaned text from Wikipedia\footnote{\url{https://en.wikipedia.org/wiki/}} to train word embeddings. For the image part, we use a subset of 16,000 images extracted from ImageNet \cite{deng2009imagenet}. Each word in WordNet \citep{miller1995wordnet} is attached with several images in ImageNet. To match the images to a word, we select the images in all the synsets where the word appears, and the images of the hyponyms of the word. We randomly split the image dataset into training set (14,000), validation set (1,000) and test set (1,000).
For each image-word pair, we randomly choose 99 other images or words (depending on the matching direction) to serve as the negative samples. The randomly sampled items have no overlap with the real set of the word or image. The model is asked to find the most relevant items among these 100 items.

\subsection{L2-Distance Baseline Model}
We propose a baseline model for the task of matching between images and words. We directly minimize the L2-loss (the euclidean distance between image vector and word embedding) to train the image projection procedure. The word embeddings of the words are pre-trained on the same corpus.

\subsection{Image to Word Vector Matching}
\begin{table}[tb]
\centering
\begin{tabular}{l|ccc}
\hline
Model & P@1 & P@5 & P@10  \\ \hline
L2-Distance Baseline & 28.8 & 55.4 & 69.3\\
IESG Model (proposal) & 62.9 & 80.1 & 85.8 \\
\hline
\end{tabular}
\caption{Image to Word vector matching results (find word given image). P@x means if one word in the top \textit{x} predicted words is correct, we think it is correct.}\label{tab:image-word match results}
\end{table}
In Table \ref{tab:image-word match results}, we show the results of Image-Word matching. The task is to find the most relevant word given an image. We assume that if the images and corresponding words can be matched based on their vectors, they are more likely to be in the same hyper-plane, which is the objective of our paper. 
From the results we can see that our IESG model can achieve very high results. P@10 even reaches 85.8\%, which means that the 85.8\% the times top 10 results contain the correct one. This shows that the image vectors can be easily connected with the correct word embeddings, indicating a strong correlation between the image vectors and the word embeddings. On the contrary, when directly minimizing the L2 distance between image vector and word embedding, the P@1 performance is much worse than our proposed model (28.8\%). We assume that this is because L2 distance model only drags the distribution of image vectors to the distribution of word embeddings while fails to make use of the context of the word.

\subsection{Word to Image Vector Matching}
\begin{table}[tb]
\centering
\begin{tabular}{l|ccc}
\hline
Task & P@1 & P@5 & P@10  \\ \hline
L2-Distance Baseline & 28.3 & 46.6 & 56.9\\
IESG Model (proposal) & 31.0 & 52.3 & 62.5 \\ \hline
\end{tabular}
\caption{Word-Image vector matching results (find image given word). P@x means if one image in the top \textit{x} predicted images is correct, we think it is correct.}\label{tab:word-image match results}
\end{table}
In Table \ref{tab:word-image match results}, we show the results of word-image matching. The task is to find the most relevant image given a word. We assume that if the images and corresponding words can be matched based on their vectors, they are more likely to be in the same hyper-plane, which is the objective of our paper. 
From the results we can see that finding the image given a word is much more difficult than the other way around. Even P@10 can only reach 62.5 which is similar to the level of P@1 for Image to Word matching. We think this is because images are much more complicated than word embeddings, images that are similar on pixels can be semantically different, we will show an example in Section \ref{sec:error analysis}. Furthermore, the compression procedure of ResNet can damage the innate structure of the image which can also hurt the matching performance. Convolutional Neural Networks classify images by focusing on one part of the image, which loses the overall information of the image.

\subsection{Word Similarity Evaluation}
\begin{table}[tb]
\centering
\begin{tabular}{l|cc}
\hline
Model & Similarity & Relatedness  \\ \hline
Skip-gram  & 0.702 & 0.604  \\
IESG & 0.680 & 0.593 \\ \hline
\end{tabular}
\caption{Word similarity evaluation on WordSim353 . IESG indicates our proposed image-enhanced skip-gram model. We use Spearman rank to evaluate the correlation. P-values are all below 1e-5.}\label{tab:wordsim results}
\end{table}
In Table \ref{tab:wordsim results} we show the word vector similarity evaluation results on WordSim353 \cite{Finkelstein2002}. 
From the results shown in Table \ref{tab:wordsim results} we can see that the correlation basically remain the same between Skip-Gram Model and our Image-Enhanced Skip-Gram Model. This indicates that our model can embody the image information into word embeddings without hurting the semantic knowledge learned from text. 
 
\subsection{Error Analysis}\label{sec:error analysis}
\begin{figure}[ht]
\centering
\includegraphics[width=2.8in]{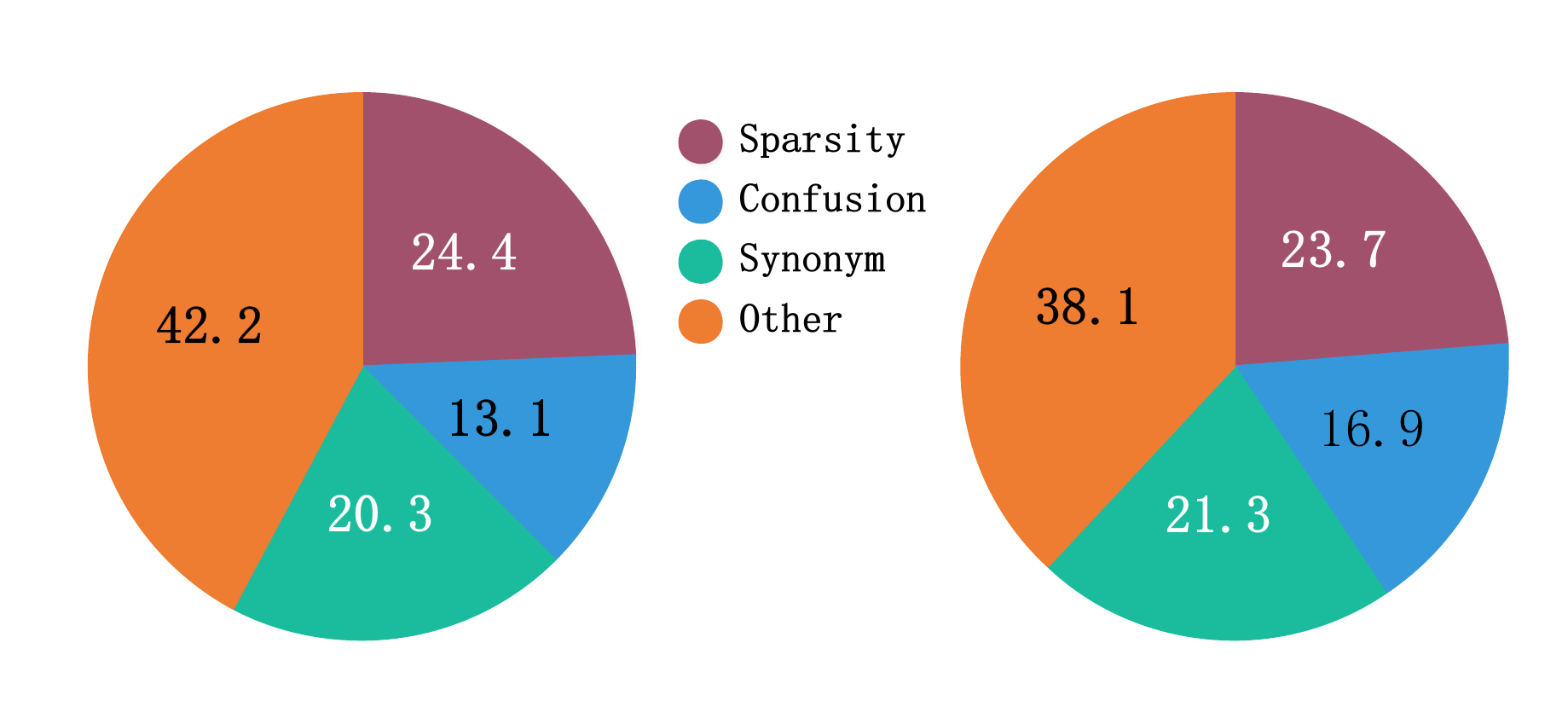}
\caption{The distribution of error types on Word-Image (left chart) and Image-Word (right chart) matching. \textit{Sparsity} means the word is very sparse in training text, \textit{Synonym} means the words are semantically similar, \textit{Confusion} means the images are similar to other images. \textit{Other} means we do not know the reason.}\label{fig:pie chart word2img}
\end{figure}
\begin{figure}[ht]
\centering
\includegraphics[width=1.8in]{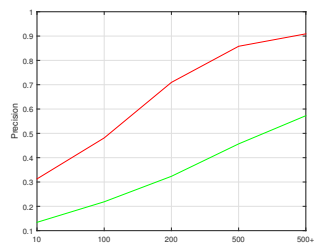}
\caption{P@1 along the term frequency of the word. Red line indicates \textit{Image-to-Word}, Green line indicates \textit{Word-to-Image}}\label{fig:tf}
\end{figure}

In Figure \ref{fig:pie chart word2img} we show the ratios of error types for \textit{Image to Word} matching and \textit{Word to Image} matching separately. 
A large part of the errors (38.29\%) in Image-to-Word matching come from \textit{Sparsity}. This can also be seen in Figure \ref{fig:tf} that the term frequency greatly affects the precision. This may come from two reasons, the entity corresponding to the word is rare in real text (e.g., triceratops); The structure of WordNet is too fine-grained (e.g., chihuahua, the distance between chihuahua and dog is 13), which makes the appearance of the word very rare when linking the image with the word. This limitation is originated from the way we collect the word-image pair.
This is also true for the ``Sparsity'' problem in the \textit{Word-to-Image} matching. To solve this kind of problem, it would be helpful to broaden the range of resources of word-image pairs and increase the number of word-image pairs. 
\begin{figure}[ht]
\centering
\begin{minipage}[t]{0.22\textwidth}
\centering
\includegraphics[width=1.5in]{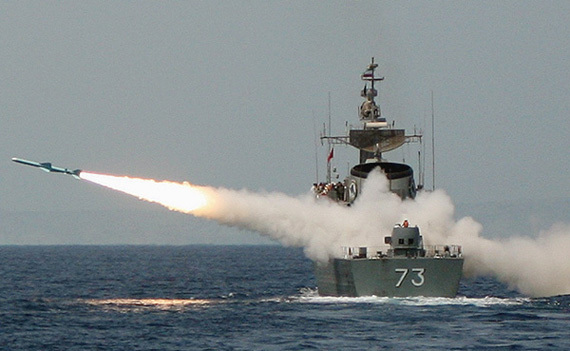}
\end{minipage}
\hfil
\begin{minipage}[t]{0.22\textwidth}
\centering
\includegraphics[width=1.3in]{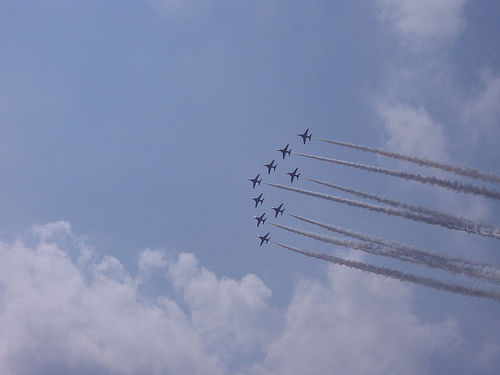}
\end{minipage}
\caption{Example of \textit{Confusing} images. \textit{Missile} (left plot) and \textit{aircraft} (right plot).}
\label{fig:confusing image.}
\centering
\begin{minipage}[t]{0.22\textwidth}
\centering
\includegraphics[width=1.5in]{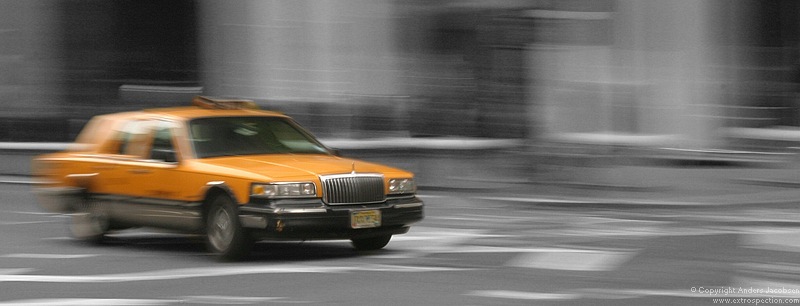}
\end{minipage}
\hfil
\begin{minipage}[t]{0.22\textwidth}
\centering
\includegraphics[width=1.3in]{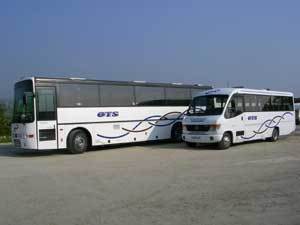}
\end{minipage}
\caption{Example of similar semantic (\textit{Synonym}) images. \textit{Taxi} (left plot) and \textit{minibus} (right plot).}
\label{fig:synonym}
\end{figure}

In Figure \ref{fig:confusing image.} we show examples of images that are confusing. In the figure, the \textit{missile} (left) is very easy to be mistakenly recognized as an ``aircraft'' (right). Actually, the top five words for the figure are ``aircraft, instrument, ammunition, weapon, missile''. These candidates are either visually similar or semantically similar to the actual answer ``missile''. We assume this error is related to ResNet, which captures a part of the figure. If the model can recognize the battleship beside the missile, the accuracy may increase. Therefore, a more appropriate model to process images is needed.

In Figure \ref{fig:synonym}, we show examples of words that are semantically similar. Given the word \textit{taxi}, the top five images are ``minibus, moped, trolleybus, projectile, triceratops''. Except for the last image (triceratops, sparsity problem), all other four images are about vehicles. We think that this is because the word embeddings are coarse-grained, bigger size of training data and a more complex language model may be helpful in this case.

\section{Conclusion}
In this paper, we explore to learn grounded word embeddings by associating the corresponding images to the word embeddings. We propose an Image-Enhanced Skip-Gram Model to jointly learn image vectors and word embeddings in the same hyper-plane. The experiment results show that the words and images are highly correlated. 

\bibliography{emnlp2018}
\bibliographystyle{acl_natbib_nourl}

\end{document}